\documentclass{article}
\usepackage{ijcai18}

\usepackage{times}
\usepackage{xcolor}
\usepackage{soul}
\usepackage[utf8]{inputenc}
\usepackage[small]{caption}
\usepackage{longtable}
\usepackage{multirow}
\usepackage{makecell}

\usepackage{amssymb}
\usepackage{amsmath}
\DeclareMathOperator{\rcos}{rcos}
\usepackage{graphicx}
\usepackage{subcaption}

\title{Elements of Effective Deep Reinforcement Learning \\
towards Tactical Driving Decision Making}

\author{Jingchu Liu, Pengfei Hou, Lisen Mu, Yinan Yu, and Chang Huang\\ 
Horizon Robotics, inc.\\
\{jingchu.liu, pengfei01.hou, lisen.mu, yinan.yu, chang.huang\}@hobot.cc
}

\begin{document}

\maketitle

\begin{abstract}
Tactical driving decision making is crucial for autonomous driving systems and has attracted considerable interest in recent years.
In this paper, we propose several practical components that can speed up deep reinforcement learning algorithms towards tactical decision making tasks:
1) non-uniform action skipping as a more stable alternative to action-repetition frame skipping,
2) a counter-based penalty for lanes on which ego vehicle has less right-of-road,
and 3) heuristic inference-time action masking for apparently undesirable actions.
We evaluate the proposed components in a realistic driving simulator and compare them with several baselines. Results show that the proposed scheme provides superior performance in terms of safety, efficiency, and comfort.
\end{abstract}

\section{Introduction}
\label{sec:intro}
Autonomous driving has attracted considerable interest in the past two decades and significant progress has been achieved.
According to Douges~\cite{Donges_1999}, autonomous driving tasks can be roughly classified into three categories: navigation, guidance, and stabilization. Strategic navigation tasks are responsible for generating road-level routes. Tactical-level guidance tasks are responsible for guiding ego vehicle along these routes in complex environments by generating tactical maneuver decisions. And operational-level stabilization tasks are responsible for translating tactical decisions into reference trajectories and then low-level control signals.
Among these three classes of tasks, tactical-level decision making is especially important due to its central role and has been an active field of research.

Early successes of decision making systems typically rely on human-designed rules to control the decision process, using methods such as heuristic rules, decision trees, finite state machines, or fuzzy-logic \cite{Montemerlo_2008,Urmson_2008,Miller_2008}. These methods are often tailored for specific environments and do not generalize robustly.

More recently, the problem of tactical decision making has been cast into the Partially Observable Markov Decision Process ({POMDP}) framework
and various approximate methods have been proposed to solve the theoretically intractable models for tactical decision making ~\cite{Ulbrich_2013,Brechtel_2014,Galceran_2015}.
One common problem faced with POMDP-based work is the strong dependency to a relatively simple environment model, usually with delicately hand-crafted (discrete) observation spaces, transition dynamics, and observation mechanisms. These strong assumptions limit the generality of these methods to more complex scenarios.

In recent years, the success of deep learning has revived the interest in end-to-end driving agent which decides low-level control directly from image inputs, using supervised learning ~\cite{Bojarski_2017} or reinforcement learning ({RL}) \cite{Sallab_2017,Plessen_2017}. But the black-box driving policies learned by these methods are susceptible to influence under drifted inputs. Although efforts have been made to identify a more robust and compact subset of prediction targets than control outputs (e.g. in \cite{Chen_2015}),
most practical autonomous driving systems to date still only use deep learning as a restricted part of the whole system.

Deep RL is a natural way to incorporate deep learning into traditional POMDP or RL-based decision making methods. 
The use of function approximators makes it possible to directly use high-dimensional raw observations. This alleviates the strong dependency to hand-crafted simple models in traditional POMDP and RL-based work.
Along this line of research, ~\cite{isele2017navigating} and ~\cite{mukadam2017tactical} apply the deep Q-network (DQN) ~\cite{Mnih_2013} to learn tactical decision policies for intersection crossing and lane changing on freeway, respectively.
Hierarchical RL is combined with Monte-Carlo tree search (MCTS) in ~\cite{Paxton_2017} to simultaneously learn a high-level option policy for decision making and a low-level policy for option execution.
~\cite{SSSS_2016,SSSS_2017} combine a high-level RL policy with a non-learnable low-level policy to balance between efficiency and safety.

However, many commonly-used techniques for deep RL are originally proposed for low-level control tasks and can be less efficient and/or robust for high-level tactical decision making from our observation.
Firstly for temporal abstraction, frame skipping with action repetition will cause unstable low-level behavior due to the discontinuity in high-level action semantics.
Secondly for the multi-dimensional rewarding systems, we find that the commonly used sparse global goal indicators and dense local goal metrics are in general redundant and harmful, and a constant value lane penalty is hard to induce favorable lane switching behavior in multi-lane scenarios.
Thirdly, decision agents relying mere on a learned RL policy may accidentally issue disastrous action under the influence of observation noise.

In this paper, we aim to tackle the above difficulties for deep RL. Our main contributions are a set of practical yet effective elements for deep RL agents towards tactical driving decision making tasks. We propose non-uniform action skipping as a more stable alternative to action repetition. A counter-based lane penalty is also proposed to encourage desired behavior for multi-lane scenarios. During inference, the learned RL agent can be further enhanced with heuristic rules that filter out obviously undesirable actions. These elements are meant to make as less modification to existing methods as possible for the sake of simplicity, and target the peculiarity of high-level tactical decision making for effectiveness. The proposed elements are equipped in a hierarchical autonomous driving system and the effectiveness is demonstrated in realistic driving scenarios presenting two-way traffic and signaled intersections.

\section{Method}
\label{sec:method}
We consider a hierarchical autonomous driving system that orchestrates learning-based and non-learning modules: the tactical decision making module is implemented as a deep RL agent for efficiency while the routing, planning, and control modules are realized with non-learning methods for safety and comfort.
The routing module calculates local lane suggestions towards the global goal according to a road map.
The decision module takes into consideration both the routing suggestions and other information such as the status of ego vehicle and road structure to make high-level maneuver directions.
The planning module then translates those directions into the trajectories of vehicle poses.
The control module finally implements the planning trajectories into low-level control signals.
Note the planning module has certain built-in safety functions to avoid hazards such as collisions.

From a agent-centric perspective, the tactical decision agent makes sequential decisions in the dynamic environment composed of all non-learning modules and the rest of the world.
In time step $t$, information about ego vehicle and the surrounding environment is compiled into an observation $o_t$ and presented to the agent.
The agent then selects a tactical decision action $a_t=\pi(o_t)$ according to the decision policy $\pi(\cdot)$.
Downstream modules will receive this decision and control the movement of ego vehicle accordingly.
A rewarding module will then assess the movement in current time step to provide a scalar reward $r_t$.
And the system will evolve forward in time into the next time step $t+1$.
The goal is to learn a policy that maximizes the expected total discounted reward
\begin{equation}
    \mathop{\mathbb{E}}\left\{G_t\right\} = \mathop{\mathbb{E}}\left\{\sum_{\tau=t}^{T}{\gamma^{\tau - t} R_\tau}\right\}.
\end{equation}
Note the world state is only partially observable to the decision agent due to imperfect sensing and unpredictable behavior of other agents. Therefore, we extend the observation vector $o_t$ into history by means of frame stacking. Other methods, e.g. recursive neural networks, can also be used to fill more information into the observation vector.

\subsection{Action Skipping}
Action-repeated frame skipping \cite{Mnih_2013} is a commonly-used technique to speed up reinforcement learning algorithms.
The benefits are multi-fold.
For exploration, non-trivial high-level maneuvers can be more easily explored with random perturbation. Otherwise the difficulty of forming a series of low-level movements that correspond to a high-level maneuver can be exponential in the length of that series.
Also, the effective horizon of the semi-MDP resulting from action repetition is proportionally shorter than the original MDP. And bootstrapped value estimation methods, such as temporal difference learning, will receive proportional speedup.
Moreover, the reward signal can become more resilient to noises and delays thanks to the extended effective period of each action.

\begin{figure}[t!]
    \centering
    \begin{subfigure}[b]{0.48\linewidth}
        \centering
        \includegraphics[width=1.0\linewidth]{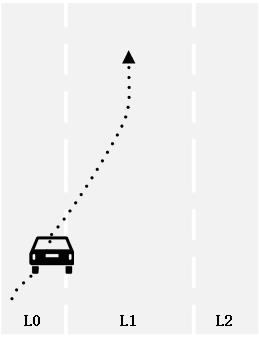}
        \caption{Before crossing lane}
    \end{subfigure}%
    ~
    \begin{subfigure}[b]{0.48\linewidth}
        \centering
        \includegraphics[width=1.0\linewidth]{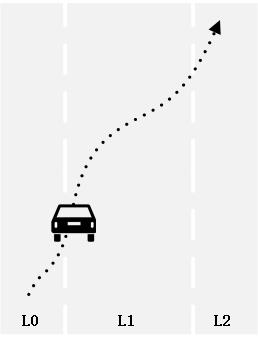}
        \caption{After crossing lane.}
    \end{subfigure}%
    \caption{Illustration for the discontinuous action semantics of lane switching tasks: the meaning of ``switching to right lane'' changes after crossing the lane marking.}
    \label{fig:lane}
\end{figure}

However, action repetition can be less stable for high-level decision making tasks due to the discontinuity in action semantics. Consider driving on a multi-lane road shown in Figure \ref{fig:lane}: when ego vehicle is just about to cross the marking between the current lane ($L_0$) and the lane immediately to the right ($L_{1}$), the action \emph{switching to the right lane} means \emph{switching from $L_0$ to $L_{1}$}. But after ego car has crossed the lane marking, the semantics of that same action is radically changed to \emph{switching from $L_{1}$ to $L_{2}$}, in which $L_{2}$ is the lane further to the right of $L_{1}$. If the frame skipping period contains such moments of non-continuous action semantics, the resulting high-level direction is doomed to result in unfavorable lower-level behaviors, e.g. frequent overshot and correction for the above scenario.

We propose to \emph{skip} instead of \emph{repeat} actions. Concretely, each agent action is now a meta action consisting of an actual decision followed by several null actions (No-Op). We denote this operation as action skipping.  During skipping, the agent can continue to collect observations and rewards. The overall reward for the meta action is then calculated as the average of all rewards collected during the skipping period. Note to implement action skipping in the decision module, the lower-level modules need to continue operation under the null action. This is not a problem as long as the trajectory planning module plans over a horizon longer than the skipping period, which is relatively easy to implement.

One drawback of action skipping is the decrease in decision frequency which will delay or prevent the agent's reaction to critical events.
To improve the situation, the actions can take on different skipping factors during inference.
For instance in lane changing tasks, the skipping factor for lane keeping can be kept short to allow for swift maneuvers while the skipping factor for lane switching can be larger so that the agent can complete lane changing actions.
When performing non-uniform action skipping, the agent may observe time phases that are skipped during training and cause domain drift in observations. As a solution, we uniformly randomly extend the skipping factor of the first agent step by a factor between zero and one so that the agent can observe all possible time phases during training.

\begin{table*}[t!]
    \centering
    \begin{tabular}{c|c|c|c}
        \bf{Category} & \bf{Name} & \bf{Description} & \bf{Weight}\\
        \hline
        \hline
        Goal indicators & N/A & N/A & N/A \\
        \hline
        \multirow{4}{*}{Constraint violation} & Collision risk & Directional risk of crashing immediate into other cars. & $-1.0$\\
          & Traffic light & Entering road section invalidated by red light. & $-1.0$ \\  \cline{2-3}
          & Dangerous lane & \makecell{Risk factor linear in the duration of \\ staying on a undesired lane.} & \makecell{Biking: $-0.2$\\ Opposite $-0.4$}\\
        \hline
        \multirow{3}{*}{Dense heuristics} & Speed & The ego car velocity along lane center. & $0.1$\\
          & Lane switching & Unit cost of switching to adjacent lanes. & $-0.4$\\
          & Step cost & Unit per-step cost. & $-0.1$ \\
    \end{tabular}
    \caption{Description for reward components and their weights.}
    \label{tab:reward_function}
\end{table*}

\subsection{Reward Function}
\label{sec:method-reward}
Tactical decision making needs to balance among efficiency, safety, and comfort. Therefore the scalar reward used for tactical decision making is usually composed of multiple components, most often through linear combination.
Roughly speaking, these reward signals can be classified into sparse global goal indicators, sparse constraint violation alerts, and dense heuristic components.
The reward components we use is shown in Table \ref{tab:reward_function} and our choices are explained below.

\subsubsection{Global Goal Indicators}
Global goal indicators are very sparse signals that only take on non-zero values when a long-term goal if achieved or missed. For tactical decision making, the true long-term goal is reaching the destination given by the navigation module as fast as possible. Therefore the most common form of global goal indicators is a signal given at the end of each episode, positive if ego car reaches the destination and negative otherwise. In this way, the discounted total reward will be larger for episodes in which ego car reach the destination earlier.

We argue that global goal indicators are not only unnecessary but also burdens to the tactical decision making agent. The preferences described by global goal indicators (i.e. what is wanted) can be implicitly expressed with some denser reciprocal constraint violation alerts (i.e. what is unwanted). This is possible because all behaviors that will stop ego vehicle from reaching the destination can be defined as violating constraints and harshly penalized. The use of dense per-step cost can also further devalue behaviors that will mislead ego vehicle into dead-end situations. As a result, the behavior that will help achieve a global goal will naturally result in low penalty. In comparison with sparse indicators, their denser counterparts will usually result in faster credit assignment during value estimation and is therefore more desirable. Moreover, global goal indicators emitted at the end of episodes will generally increase the absolute value of expected accumulated rewards in comparison with the indicator-free counterpart. In turn, the approximators used for value functions needs to have larger dynamic range, which mean more learning steps during training and larger variance during inference. For these reasons we do not use any global goal indicator components in our experiments.

\subsubsection{Constraint Violation Alerts}
Constraint violation alerts are sparse signals that penalize the agent for being in a risky situation or performing a risky action. The most common situation considered is collision with other road objects, e.g. other vehicles or pedestrian. Note reward signals that fall into this category need not to be sparse during constraint violations events. They are sparse in the sense that risky situation should be very rare under a properly cautious driving policy.

We consider three types of risky situations in our experiments: 1) entering intersection during red light, 2) collision with other vehicles, and 3) treading onto biking and opposite lanes on which ego vehicle has less priority. The former two components will also termination the episode.

\paragraph{Traffic light:} The traffic light alert is triggered when ego vehicle enters an intersection when the corresponding connection road is covered by a red light. Note although we render longitudinal speed control to a rule-based planning module and it will automatically stop ego vehicle in most situations, there are still corner cases that may accidentally grant ego vehicle's access into intersection during red lights. Therefore a traffic light alert should be in place to penalize these corner-case behaviors.

\paragraph{Collision risk:} The collision risk component is active when ego vehicle is about to crash into other vehicles. It is the sum of risk components contributed by each of the other vehicles in the region of interest. Each component is further defined as the product of an isotropic distance-based factor and a directional factor related to the heading direction of cars:
\begin{equation}
    r_\textrm{c} = \sum_i{r_\textrm{u}^i \cdot r_\textrm{d}^i},
\end{equation}
where $r_\textrm{u}^i$ is the isotropic factor by target $i$ and $r_\textrm{d}^i$ is the directional factor.
The distance-based risk factor is Laplacian in the distance between the target car and ego car:
\begin{equation}
    r_\textrm{u}^i(d) = e^{-d+d_0},
\end{equation}
where $r_\textrm{u}^i(d)$ is the distance-based risk for target car $i$ at distance $d$ and $d_0$ is a normalizing distance.
The directional risk factor is the product of two narrow-band raised-cosine patterns, the center of which are aligned with the heading direction of ego car and the target car, respectively:
\begin{equation}
    r_\textrm{d}^i(\theta_{\textrm{ego}}, \theta_{\textrm{target}}) = 
       \rcos(\theta - \theta_{\textrm{ego}}) \cdot
       \rcos(\theta - \theta_{\textrm{target}}),
\end{equation}
where $\rcos(\cdot)$ is the narrow-band raised cosine function, $\theta_{\textrm{ego}}$ and $\theta_{\textrm{target}}$ are the heading angle of ego and target vehicle, and $\theta$ is the direction of the vector connecting ego car and target car. The overall effect of these two factors is high risk only when ego car and a target car is about to drive head-to-head into each other. Otherwise the risk is relatively low, e.g. when ego car and the target car is driving closely side-by-side. 

Another commonly-used measure for crashing risk is Time-to-Collision ({TTC}). TTC is roughly inversely proportional to the distance between ego and the target car. Therefore it is relatively aggressive in risk prediction and tend to exaggerate crashing risk. In contrast, the proposed risk formulation indicates risk only when it is about to happen and is thus more conservative. To foresee upcoming hazards, the agent can use common value estimation methods to predict future risk values.

\paragraph{Dangerous lane:}
We use counter-based risk signals to reflect the ever-increasing empirical risk of staying on biking and opposite lanes. Specifically, the indicator for each lane maintains a counter (with a maximum value cap) that keeps track of the time steps that ego vehicle has spent on that lane. The risk value is then computed as a linear function of the corresponding counter value:
\begin{equation}
    R = (0.1 x + 0.9) \times (x > 1.0),
\end{equation}
where $R$ is the risk value and x is the counter value.
The lane risk defined as such will be relatively small when ego vehicle just arrived on a dangerous lane and will gradually become intolerably large if ego vehicle remains there for a long time. In this way, temporally switching onto dangerous lanes, which is required for overtaking slower traffic on single lane roads, can be enabled. And staying on dangerous lane will be prohibited in the long run. Note since the reward component defined in this way becomes stateful, it is important to augment agent observations with history information so that it can roughly infer how long it has stayed on a particular lane.

In contrast, it is much more difficult to design constant-value risk signals that has the same enabling effect on overtake maneuvers: a small risk value may encourage ego vehicle to switch onto dangerous lanes when necessary, but it may easily fail on encouraging backward lane switching because the small difference in risk value can be easily overwhelmed by the variance in approximated value functions. The agent can only slowly learn the risk of dangerous lanes from sparse collision events. Meanwhile, a large risk value will in effect prohibit switching onto dangerous lanes, even when doing so is beneficial.


\subsubsection{Dense heuristic metrics}
Reward signals belonging to this category are usually used to hard-code designer's heuristic preference for some states and actions.  Unlike the former two reward categories, which are easier to design as they can reflect orthogonal aspects of desired and unwanted outcomes, dense heuristic metrics are harder to design because heuristic rules can easily conflict with each other or, even worse, fail to align with the global goals. For this reason, we aim to employ only a minimal set of dense heuristic components.

We consider a component proportional to the speed of ego vehicle along the navigation route to encourage driving towards the goal as fast as possible. The speed limits are monitored by the planning module to avoid over-speeding. The second dense component we consider is a per-step penalty for lane changing actions to discourage unnecessary lateral maneuvers and improve passenger comfort. The final dense component applied is a trivial per-step cost to prefer short episodes. We do not employ any dense penalties related to local goals (e.g. local target lane or headway to other vehicles) as they can easily conflict with other heuristic metrics and the global goal.

\subsection{Rule-based Action Masking}
In some scenarios, undesirable tactical actions can be straightforwardly identified. In such cases, we proposed to apply simple rules to filter out those actions during inference instead of only hoping the agent to learn to avoid those actions. The reason is that, on one hand, even if the agent can learn to avoid inferior actions, they can still be triggered due to the variance in observation and the learned model. On the other hand, those simple rules designed for straightforward situations are less prone to unexpected false positives and easier to debug if any happens. This is in contrast to conventional rule-based decision policies which is comprised of a complex set of rules intended to work under complex scenarios.
\section{Implementation}
\label{sec:implementation}

\begin{figure}[t!]
    \centering
    \begin{subfigure}[b]{0.48\linewidth}
        \centering
        \includegraphics[width=1.0\linewidth]{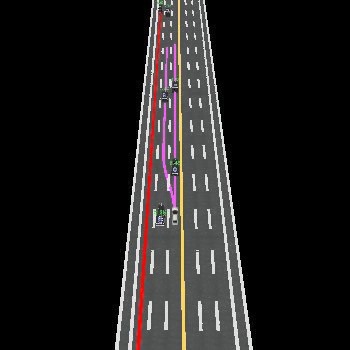}
        \caption{ }
        \label{fig:a}
    \end{subfigure}%
    ~
    \begin{subfigure}[b]{0.48\linewidth}
        \centering
        \includegraphics[width=1.0\linewidth]{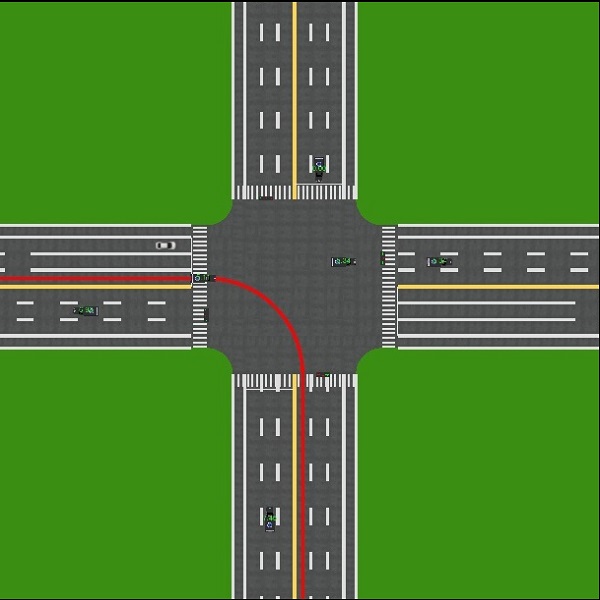}
        \caption{ }
        \label{fig:b}
    \end{subfigure}%
    \caption{(a) Illustration of the tilted top-down view used for agent observation. (b)  Simulation scenario with signaled intersection and two-way traffic. Red line indicates the navigation route.}
    \label{fig:ab}
\end{figure}

As shown in Figure \ref{fig:a}, the decision agent observes tilted RGB top-down views of ego vehicle's immediate surrounding.
In each time step, the latest two frames from the rendered $10$Hz image stream are max-pooled pixel-wise into a single image frame to combat flickering. The frames from the latest three time steps are further stacked channel-wise to form a single observation.
The reward components and linear combination weights defined in Section \ref{sec:method-reward} are used to derive a scalar reward function. For comparison, some of the components may be removed or replaced for comparison.

The agent is composed of a dueling deep Q-network (DQN) ~\cite{dueling_2015}, with $3$ convolutional layers followed by $2$ dense layers and one linear output layer. The last dense layer is divided into an action-wise advantage channel and a baseline channel. The {Q} value for each action is the sum of its corresponding advantage plus the shared baseline.
The three convolutional layers has $16$, $16$, and $64$ kernels with size $8$, $5$, $3$ and stride $4$, $3$, and $2$, respectively. The first dense layer has $256$ hidden units while the latter dense layer has $256$ hidden units for each of the two channels.
All convolutional layers apply $3 \times 3$ max pooling and all hidden layers apply {ReLU} activation.

The discount factor used is $0.9$ and double Q learning \cite{doubleq_2015} is applied to build up the temporal difference loss.
We use a mini-batch of $8$ samples and an ADAM optimizer with learning rate $1e^{-4}$, $\beta_1 = 0.9$, and $\beta_2 = 0.999$ to train the learning network.
The target network uses a synchronization rate of $1e^{-3}$ to track the learning network.
The exploration factor is annealed linearly from $0.2$ to $0.05$ in $30$K steps.
The training process is handled asynchronously by a rate-controlled training thread: for each data point collected from the environment, this thread can perform $8$ updates for the learning network.
The training thread samples from a memory-mapped on-disk replay buffer of maximum size $300$K. The data is divided into consecutive fragments of size $100$ and about $30$ fragments are cached in memory for random sampling at any time. Each fragment will be sampled for at most $200$ times and a new fragment will be swapped in for replacement.

\section{Experimental Results}
\label{sec:sim}
\begin{table*}[]
    \centering
    \begin{tabular}{c|c|c|c|c||c|c|c}
\bf{Section}  &\bf{ID} &  \bf{Skipping}  & \bf{Reward}   & \bf{Other}     & \bf{Success Rate}   & \bf{Lon. Speed}   & \bf{Lat. Speed}    \\
\hline
\hline
\multirow{2}{*}{Baseline}  &01      & \multirow{2}{*}{N/A}     &\multirow{2}{*}{N/A}      &   Random                            &   0.080              &    4.85              &   0.496   \\
                           &02      &                          &                          &   Rule-based                        &   0.790              &    5.11              &     0.0822      \\
\hline
\multirow{3}{*}{Skipping}  &03      &None                      &\multirow{3}{*}{Proposed} &  \multirow{2}{*}{N/A}               &   0.295              &    3.49              &     0.00538     \\
                           &04      &Repetition                &                          &                                     &   0.840              &    5.64              &     0.204  \\ \cline{5-5}
                           &05      &Dynamic                   &                          &   $2\times$ / $6\times$ skipping    &   0.275              &    4.96              &     0.182  \\

\hline
\multirow{4}{*}{Reward}    &06      &\multirow{4}{*}{Uniform}  &   Global goal            &  \multirow{4}{*}{N/A}               &   0.825              &    5.80              &     0.148   \\
                           &07      &                          &   $1.0$ lane penalty     &                                     &   0.820              &    5.50              &     0.0893   \\
                           &08      &                          &   $0.1$ lane penalty     &                                     &   0.295              &    4.17              &     0.0385  \\
                           &09      &                          &   Local goal             &                                     &   0.855              &    5.52              &     0.102 \\
\hline
\multirow{5}{*}{Proposed}  &10      &\multirow{3}{*}{Uniform}  & \multirow{5}{*}{Proposed}&   N/A                               &   0.830              &    5.97              &     0.162 \\
                           &11      &                          &                          &   Action mask 1                     &   0.845              &    6.05              &     0.159 \\
                           &12      &                          &                          &   Action mask 2                     &   0.900              &    5.88              &     0.131 \\\cline{3-3}
                           &13      &\multirow{2}{*}{Non-Uniform}  &                      &   Action mask 1                     &   0.860              &    5.95              &     0.191 \\
                           &14      &                          &                          &   Action mask 2                     &   0.905              &    6.11              &     0.153  
    \end{tabular}
    \caption{Experimental results.}
    \label{tab:results}
\end{table*}

As shown in Fig.\ref{fig:b}, we experiment with different agent configurations in simulated driving scenarios with two-way traffic and signaled intersections. 
The simulator is wrapped as an RL environment complying to the OpenAI Gym API ~\cite{1606.01540} and the frequency of agent-environment interaction is regularized to $2$Hz in terms of simulation clock.

Agents are trained and tested with episodic simulation.
In each episode, a route is sampled randomly from the road map of $20$ routes. Each route consists of two road segments connected by an intersection.
A number of vehicles are scattered along this route with random starting points and cruising speed.
Ego vehicle will then start from the beginning of the selected route.
A global navigation module will constantly provide a reference lane (rendered with a red line in observed image) that leads to the route destination.
An episode is terminated if ego vehicle encounters any of the following conditions: 1) reaching destination; 2) receiving a crashing risk value $>1.0$; 3) stop moving for $40$ seconds\footnote{This happens when ego vehicle drives into dead-end lanes.}; 4) entering intersection on unpermitted lanes.
Each agent configuration is trained from scratch for $10$ simulation runs. Each run terminates once the accumulated number of environment steps exceeds $250$K. The trained agents are then evaluated without exploration against $100$ pre-generated test episodes. The number of other vehicles present in each episode is $32$ for both training and test.

Test results are shown in Table \ref{tab:results} and organized into $4$ sections. We select success rate, longitudinal speed, and lateral speed as the performance metrics to reflect safety, efficiency, and comfort. The agent needs to switch to the correct lane and avoid collisions to successfully finish each episode. Overtaking slower traffic is also required to achieve a high longitudinal speed. Moreover, the agent needs to avoid unnecessary lateral maneuvers in order to reduce lateral speed. The overall metric for each configuration is calculated as the median per-step metric of all $10$ simulation runs. 

The first section presents the random and rule-based baselines. The simulation scenario is complex enough such that a random policy can only achieve a success rate of $8\%$. And the rule-based decision agent can improve the success rate metric to a more reasonable number of $79.0\%$. The longitudinal speed of rule-based of agent is higher than the random agent thanks to overtake maneuvers. And the lateral speed is reduced because a reasonable decision policy will avoid lane changing and stay on the current lane for most of the time.

The second section compares different skipping configurations. The RL agent without any skipping operation (ID$03$) performs rather poor, achieving a success rate of only $29.5\%$. Action repetition operations (ID$04$) can help significantly improve the success rate to $83\%$. This demonstrates the importance of temporal abstraction to high-level tactical decision making tasks. The success rate of dynamic action skipping scheme (ID$05$) is surprisingly unsatisfactory even though its action space is a super-set to the action repetition scheme.  One possible explanation may be that the gains from broadening the action space is over-weighed by the disadvantage of extending the effective horizon . As expected, action skipping (ID$10$) will result in smaller lateral speed than action repetition (ID$04$) thanks to the elimination of overshoot-correction jitters while roughly preserving the success rate.

The third section compares alternatives rewarding schemes to the proposed one. In comparison to experiment ID$10$, experiment ID$06$ superimpose a $\pm 1$ global goal indicator to the proposed reward function and cause both success rate and longitudinal speed metrics to deteriorate. This corroborates our previous statement that global goal indicators are redundant under the presence of a complete set of constraint violation penalties. Experiment ID$07$ and ID$08$ replaces the proposed counter-based penalty with a constant penalty. The resulting agent behavior is unfavorably sensitive to the penalty value: a larger value completely bans the access to dangerous lanes, resulting in lower longitudinal speed than experiment ID$10$; while a smaller value fails to reflect the risk of those lanes, resulting in a drastically reduced success rate. Experiment ID$09$ adds a dense penalty for deviating from the navigation lane. Although ego vehicle can achieve a higher success rate by sticking to the navigation lane, it can also miss the opportunity to temporarily deviate from that lane and overtake slower traffic. The result is significantly lower longitudinal speed. This exemplifies how dense local-goal related rewards can conflict from the global goal.

The effectiveness of rule-based action masking and non-uniform action skipping is illustrated in the fourth table section. We experiment with two set of rules: rule \#$1$ filters out lane switching behavior while ego vehicle is moving slowly on the navigation lane, while rule \#$2$ also banns ego vehicle from treading onto the opposite lane or biking lane in addition. Note rule \#$2$ is stricter than rule \#$1$ but also requires more structural information about the environment. A trade-off between safety and efficiency can be identified by comparing Experiment ID$11$ and ID$12$: the stricter rule \#$2$ can provide more significant improvements on success rate than rule \#$1$, at the price of reduced longitudinal speed.
Finally, observe from the last two experiments, non-uniform action skipping can further increase the success rate of both masking rules by giving the ego car more lane-changing opportunities during inference.


\section{Related Work}
\label{sec:related}
Temporal abstraction is an effective means for speeding up RL algorithms.
Although frame skipping with action repetition~\cite{Mnih_2013} is a very simple form of temporal abstraction, it is extremely effective and has been shown to be key to state-of-art performance in many low-level control tasks~\cite{BHMM_2000}. Due to the discontinuous semantics of tactical driving decisions, we  propose to replace action repetition with the more stable action skipping method. Dynamic frame skipping~\cite{LSR_2017} is also investigated in~\cite{isele2017navigating} for high-level decision learning, but it is not as effective as action skipping in our experiments. Other more sophisticated form of temporal abstractions have also been proposed for RL-based decision making under the option framework~\cite{SPS_1999}, e.g. \cite{SSSS_2016}. But they are mostly tailored for specific scenarios and hard to generalize.

An appropriate multi-dimensional rewarding system is indispensable for tactical decision making agents that are based on RL or POMDP. Many existing work applies global goal indicators, e.g.~\cite{Brechtel_2014,isele2017navigating,mukadam2017tactical,Paxton_2017}. However, we show that it is unnecessary to use global goal indicators in tactical decision making tasks. Constraint violation alerts in existing work are mostly one-shot or constant in time~\cite{Brechtel_2014,isele2017navigating,Paxton_2017,Li_2017}. We propose to use counter-based risk signals to speed up learning in multi-lane scenarios. Dense local-goal penalties are also used in some previous work to regulate driving policy heuristically~\cite{Ulbrich_2013,Galceran_2015,Paxton_2017}. We show that such penalties can easily mislead the agent towards sub-optimal policies and should be avoided if possible.



\section{Conclusion}
\label{sec:conclusion}
Deep reinforcement learning is a promising framework to tackle the tactical decision making tasks in autonomous driving systems.
In this paper, we propose several practical ingredients for efficient deep reinforcement learning algorithms towards tactical decision making.
We propose action skipping as a more stable alternative for the commonly used action repetition scheme and 
investigate a necessary set of reward components that will guide decision making agent to learn effectively in complex traffic environments.
For more reliable inference, a heuristic rule-based action masker is combine with the learned agent to filters out apparently unsafe actions.
The proposed ingredients is evaluated in a realistic driving simulator and results show that they outperform various baseline and alternative agent configurations.

\bibliographystyle{named}
\bibliography{ijcai18}

\end{document}